\documentclass{article}
\usepackage{ijcai_26}

\usepackage{times}
\usepackage{soul}
\usepackage{url}
\usepackage[hidelinks]{hyperref}
\usepackage[utf8]{inputenc}
\usepackage[small]{caption}
\usepackage{graphicx}
\usepackage{enumerate}
\usepackage{makecell, multirow}
\usepackage{subcaption}
\usepackage{amsmath, amssymb}
\usepackage{amsthm,amsfonts}
\usepackage{booktabs}
\usepackage{xcolor} % 导入颜色包
\usepackage{algorithm}
\usepackage{algorithmic}
\usepackage{bm}
\usepackage[switch]{lineno}

\title{Toward High-Fidelity Visual Reconstruction: From EEG-Based Conditioned Generation to Joint-Modal Guided Rebuilding}

\author{
    Zhijian Gong{$^{1,2}$}, Tianren Yao{$^1$}, Wenjia Dong{$^{1,2}$}, Xueyuan Xu{$^{1,2}$}\thanks{Corresponding author}
    \affiliations
    {$^1$}Beijing University of Technology, Beijing \\
    {$^2$}Beijing Key Laboratory of Computational Intelligence and Intelligent System, Beijing
}

\begin{document}

\maketitle

\begin{abstract}
Human visual reconstruction aims to reconstruct fine-grained visual stimuli based on subject-provided descriptions and corresponding neural signals. As a widely adopted modality, Electroencephalography (EEG) captures rich visual cognition information, encompassing complex spatial relationships and chromatic details within scenes. However, current approaches are deeply coupled with an alignment framework that forces EEG features to align with text or image semantic representation. The dependency may condense the rich spatial and chromatic details in EEG that achieved mere conditioned image generation rather than high-fidelity visual reconstruction. To address this limitation, we propose a novel \textbf{Joint-Modal Visual Reconstruction (JMVR)} framework. It treats EEG and text as independent modalities for joint learning to preserve EEG-specific information for reconstruction. It further employs a multi-scale EEG encoding strategy to capture both fine- and coarse-grained features, alongside image augmentation to enhance the recovery of perceptual details. Extensive experiments on the THINGS-EEG dataset demonstrate that JMVR achieves SOTA performance against six baseline methods, specifically exhibiting superior capabilities in modeling spatial structure and chromatic fidelity.

\end{abstract}

\section{Introduction}

Reconstructing scenes based on subject descriptions and neural activity presents a challenging but promising research direction. The task not only promotes understanding into human visual cognition, but also has the potential for augmentative communication for individuals with motor or language impaired \cite{intro1health,dm-re2i}. Among various neural signals, Electroencephalography (EEG) is widely used in recent visual cognition research due to its high temporal resolution and ease of acquisition \cite{intro3EEGtime,intro4EEGfeasible}. EEG contains rich information on spatial structure cognition and color perception induced by visual stimuli which provides a foundation for high-quality visual reconstruction \cite{intro2EEGgood,cognitioncapturer}.

\begin{figure}
    \centering
    \includegraphics[width=\linewidth]{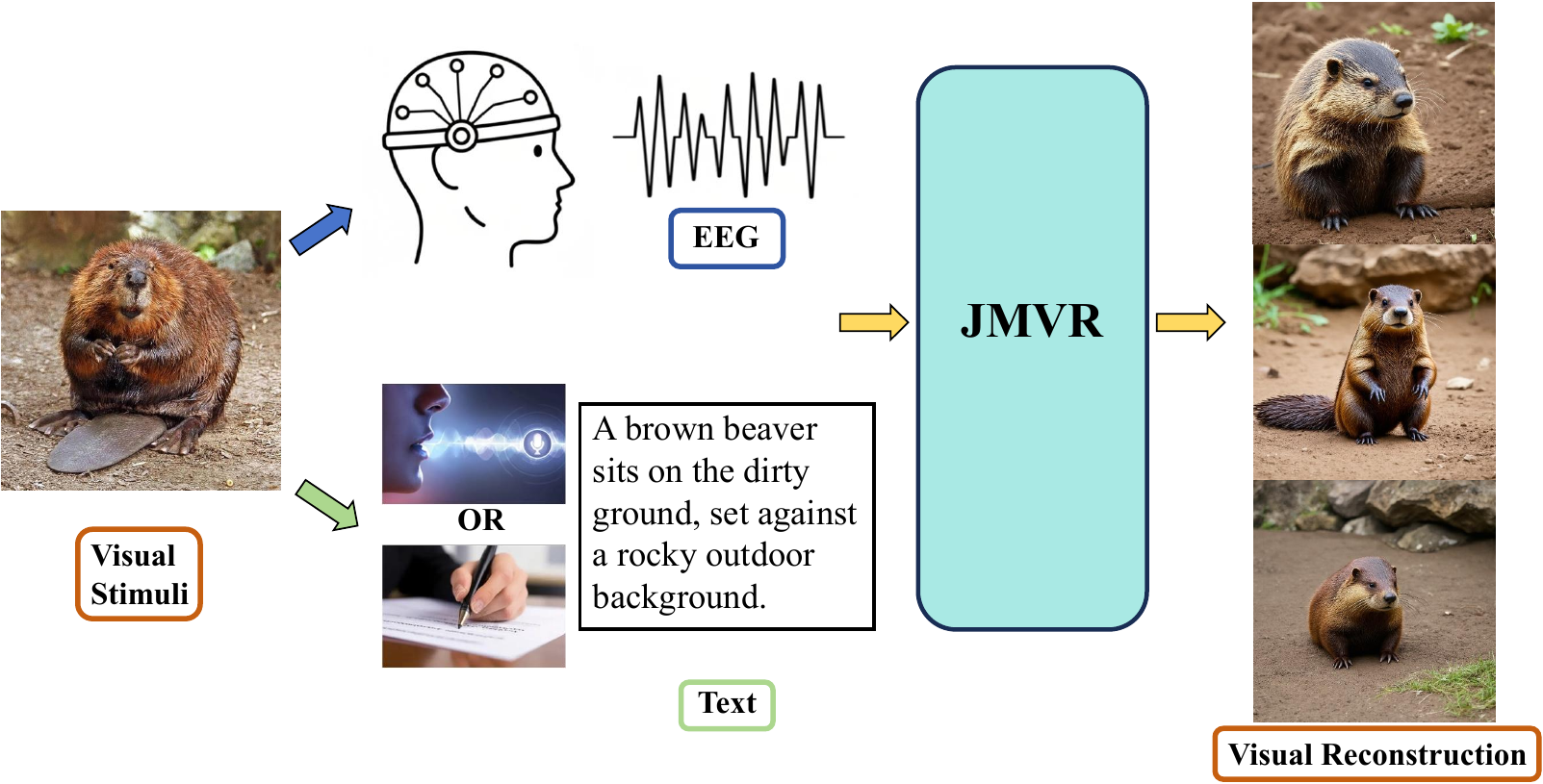}
    \caption{Overall Framework. The EEG signals and corresponding textual descriptions are input to the joint latent space in JMVR for training and visual reconstruction.}
    \label{fig:Overall}
\end{figure}

\begin{figure*}
    \centering
    % 左子图
    \begin{subfigure}[b]{0.68\textwidth}
        \centering
        \includegraphics[width=\linewidth]{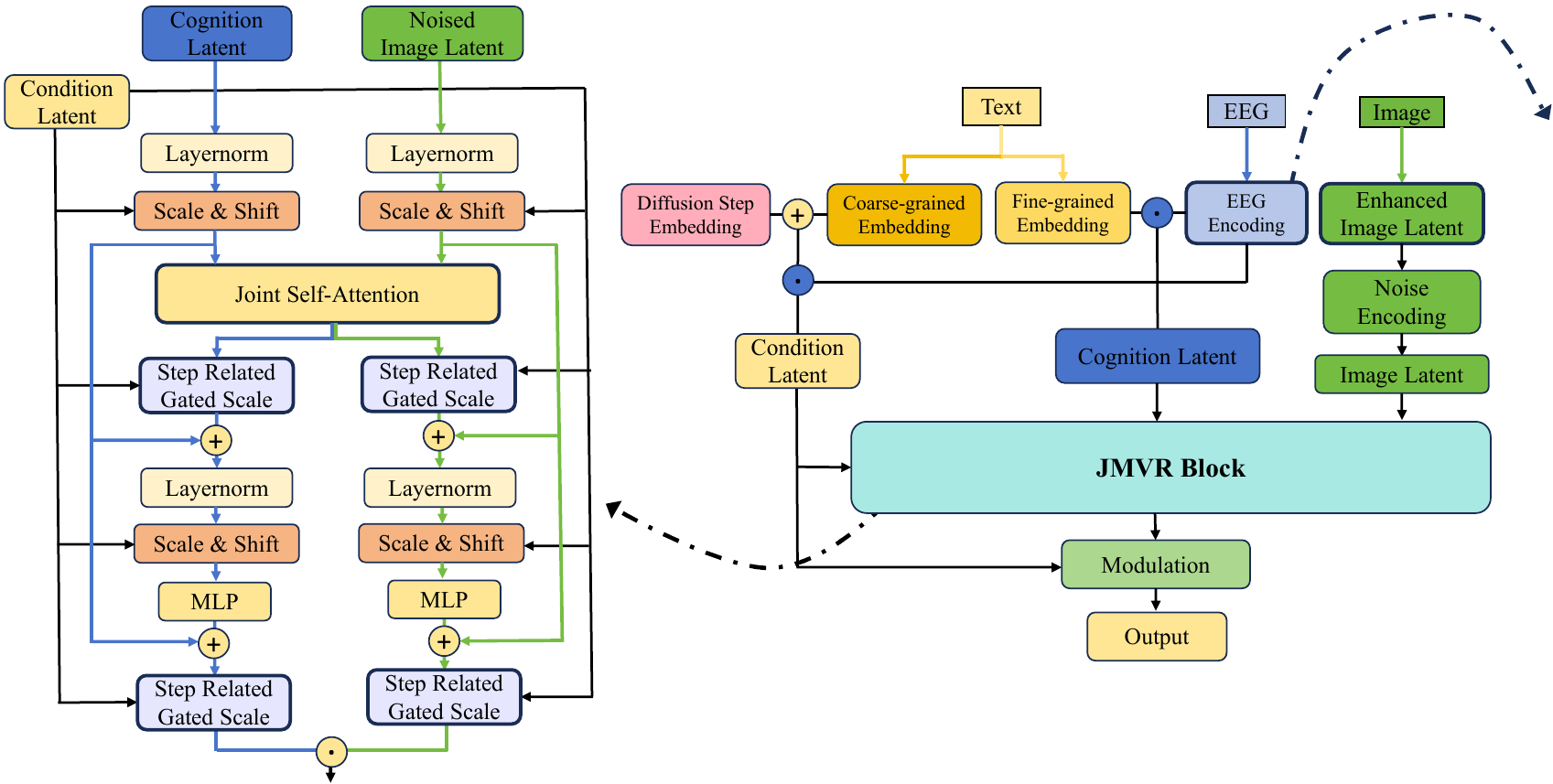}
        \label{subfig:EVCR_left}
    \end{subfigure}
    \hfill
    % 右子图
    \begin{subfigure}[b]{0.30\textwidth}
        \centering
        \includegraphics[width=\linewidth]{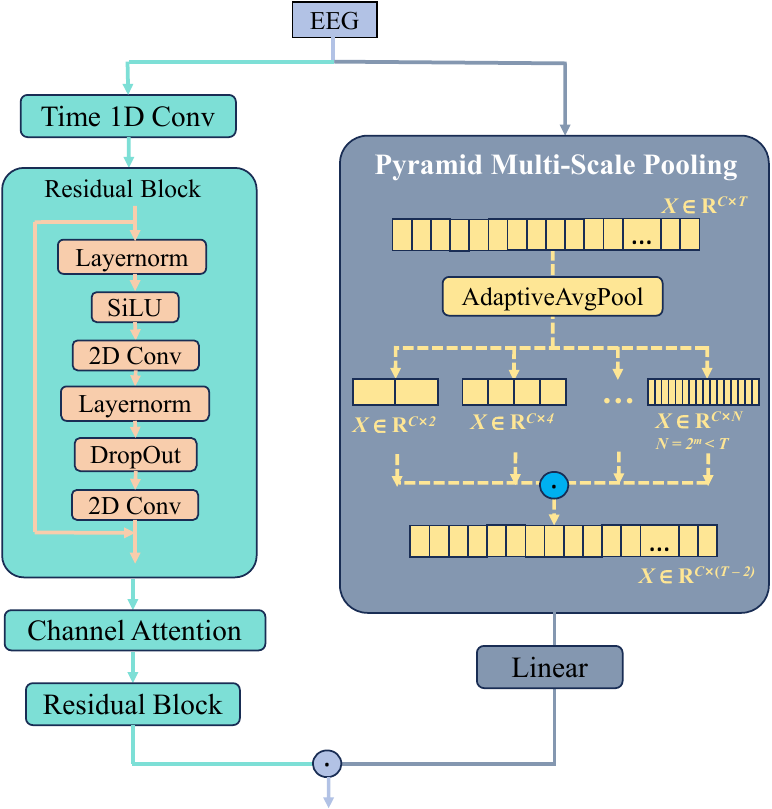}
        \label{subfig:EVCR_right}
    \end{subfigure}
    \caption{Overview of the JMVR framework. Text prompts are processed into coarse- and fine-grained embeddings. The former combines with diffusion timesteps and EEG to construct the condition latent for feature modulation, while the latter is concatenated with EEG representations to generate the cognition latent. Within the JMVR Block, this cognition latent interacts with the enhanced image latent via a joint-attention mechanism to synthesize the final output after modulation. } 
    \label{fig:JMVR_group}
\end{figure*}

Previous works have demonstrated the feasibility of decoding visual information from EEG. Early studies examined approaches based on Generative Adversarial Networks (GANs) \cite{GAN1,GAN2}, but achieved limited image reconstruction quality \cite{dm-re2i}. Recently, Denoising Diffusion Probabilistic Models (DDPMs) have been proposed to establish the foundation for high-quality conditional image generation \cite{necomimi,braindreamer,synapse}. This motivates researchers to employ EEG signals to condition DDPMs for effective image synthesis. Due to the inherently low signal-to-noise ratio characteristics of EEG, these methods typically pre-align EEG features with image or text modalities. Such text-like or image-like aligned embeddings enable a more effective capture of semantics in EEG and can be directly integrated into mature text-to-image pipelines. \cite{dreamdiffusion}.

However, human cognitive responses to visual stimuli are inherently diverse and multifaceted \cite{cognitioncapturer}. This alignment-centric paradigm may restrict the model’s capacity to reconstruct richer perceptual information, such as spatial layout and chromatic cognition \cite{clipbad}. Although existing models achieve meaningful EEG-conditioned image generation, their deep coupling with text or image modalities limits the detailed reconstruction of visual perceptions that fall outside the pre-aligned semantic space.

To address these issues, we introduce Joint-Modal Visual Reconstruction (JMVR), a novel framework that treats EEG as an independent modality without prior alignment. The key contributions are as follows: 

\begin{enumerate}[\textbf{·}]
    \item \textbf{Representation Decoupling: }JMVR decouples EEG representations from the abstract semantic of text and images. By unifying all modalities within a joint latent space, it establishes a direct mapping from EEG signals to complex visual features that prevents information degradation.
    \item \textbf{Diffusion-step Gated Scaling: }Adaptive weights are implemented to text and EEG modalities across different diffusion steps. It facilitates the model to focus on high-level textual semantics during coarse generation, while attending to visual details contained in the EEG signals in fine-grained synthesis.
    \item \textbf{Multi-scale EEG Encoding: }A multi-scale pyramid EEG encoding architecture is proposed by modeling both coarse- and fine-grained visual cognition simultaneously. This design significantly enhances the encoder’s ability for representing multi-scale visual information from EEG signals.
    \item \textbf{Image Augmentation: }Edge, saturation, and depth information are extracted from the original visual stimuli. Integrating these features into the joint latent space facilitates the model to capture perceptual details from EEG that are often lost during alignment with textual descriptions.
\end{enumerate}

\section{Related Work}
Early exploration in EEG-based visual reconstruction primarily relied on GAN \cite{GAN1,GAN2,GAN3,GAN4}. However, these approaches often limited to generate high-fidelity images \cite{dm-re2i}. Current research has pivoted toward Diffusion models to achieve better reconstruction quality. For example, NICE introduced a spatiotemporal encoder incorporating self-attention and graph attention mechanisms, while MUSE adopted a similar strategy utilizing spatial convolutions and Transformer-like graph attention to encode spatially representative EEG signals \cite{NICE,MUSE}. DreamDiffusion used an pre-trained large EEG model to replace the encoder for EEG representation \cite{dreamdiffusion}. These frameworks essentially align the entire EEG recording with images via cosine similarity to condition the generation process. As the demand for reconstruction precision has intensified, recent research has shifted toward disentangling EEG semantics through more granular alignment strategies. For instance, ATM implemented a dual-pipeline framework that aligns EEG separately with abstract textual semantics and image details \cite{ATM}. Building on this, Perceptogram explored a tripartite alignment framework that integrates text, image, and image latent features to better recover low-level attributes such as color and shape \cite{perceptogram}. Despite these advancements, the reconstruction of fine-grained visual details remains a persistent challenge. Addressing this, CognitionCapturer proposed that simple semantic alignment fails to decode visual cognition beyond high-level concepts in text and image \cite{cognitioncapturer}. It introduces a scalable multi-modal approach that incorporates auxiliary information in EEG like depth. However, this paradigm remains constrained by the alignment framework that relies on the manual provision of supplementary modalities. The lack of capacity to adaptively capture intrinsic cognitive features remains challenge.

\section{The Proposed Framework}

In this section, we present the JMVR framework that designed to reconstruct high-fidelity visual stimuli from EEG signals. As illustrated in Fig.\ref{fig:JMVR_group}, the proposed architecture treats EEG as an independent conditional modality, avoiding the information loss associated with forced semantic alignment. This framework operates on a Diffusion Transformer (DiT) backbone that acquires text and EEG signals to learn the relationship between these signals and the image denoising process in the multi-modal space \cite{DiT}. Four novel modules will be introduced in this section including Multi-scale EEG Encoder and Image Augmentation for input signals; Joint-Modal Attention and Diffusion Step Gating for joint training and modality information balance. To avoid ambiguity between the temporal dimension of EEG signals and the diffusion process, we denote the EEG signal length as $T$ and the diffusion timestep as $\tau$.

\subsection{Multi-Scale EEG Encoder}
To extract both fine-grained details and global semantics for visual reconstruction, we propose a dual-stream EEG encoder to process raw EEG signals $\mathbf{X} \in \mathbb{R}^{C \times T}$. The encoder consists of two branches that designed for capturing local spatiotemporal dynamics and holistic contextual representations. 
(1) The spatiotemporal stream on left branch focuses on modeling high-frequency EEG variations that are critical for rebuilding detailed visual information. Specifically, a 1D temporal convolution is applied to extract temporal features and a channel attention mechanism that enhances visually responsive electrodes and suppresses noise. Each residual block is designed to fuse spatiotemporal information while stabilizing training \cite{dm-re2i}. It includes  Layer Normalization, SiLU activation, and 2D convolutions. 
(2) The right branch aim to capture global information through a Pyramid Multi-Scale Pooling (PMSP) module. To complement the limited receptive field in spatiotemporal stream, PMSP leverages adaptive average pooling to construct hierarchical representations at multiple temporal scales. Specifically, pooled features with resolutions $\mathbb{R}^{C \times 2^1}, \mathbb{R}^{C \times 2^2}, \dots, \mathbb{R}^{C \times 2^m}$ are generated, where $2^m < T$. These multi-scale features are concatenated as $\mathbf{\hat{X}} \in \mathbb{R}^{C \times (T-2)}$ along the temporal dimension and subsequently projected to enable effective modeling of visual semantics from coarse to fine granularity. Finally, the outputs of the two branches are concatenated to form a unified representation that balances global visual cognition with fine-detail perception.  Further information on dimensional details are provided in the appendix.

\subsection{Image Augmentation}
To construct a semantically enriched joint latent space encapsulating multi-dimensional visual perception, we propose a Visual Feature Augmentation strategy. By explicitly integrating edge, saturation, and depth information of original visual stimuli, this approach facilitates a profound correspondence between EEG signals and visual attributes. To reinforce the modeling of structural geometry, we apply the Canny operator to generate a binary edge map after the Gaussian smoothing with a $3\times3$ kernel to the grayscale image to suppress textural noise. The edge detection threshold is set to 50$/$150. Recognizing chromatic as a significant element of visual cognition, we transform origin images into the HSV color space and extract the saturation channel. This provides the joint space with a dedicated representation of color intensity and distribution. Furthermore, to capture the neural correlates of 3D spatial structure within the latent space, we utilize the Depth-Anything-v2 model to generate high-fidelity depth maps \cite{depth_anything_v2}. Finally, the original image and these corresponding augmented views are individually encoded via a pre-trained Variational Autoencoder (VAE) and projected together through a linear layer into a unified latent embedding \cite{DiT}.

\begin{figure}
    \centering
    \includegraphics[width=0.4\linewidth]{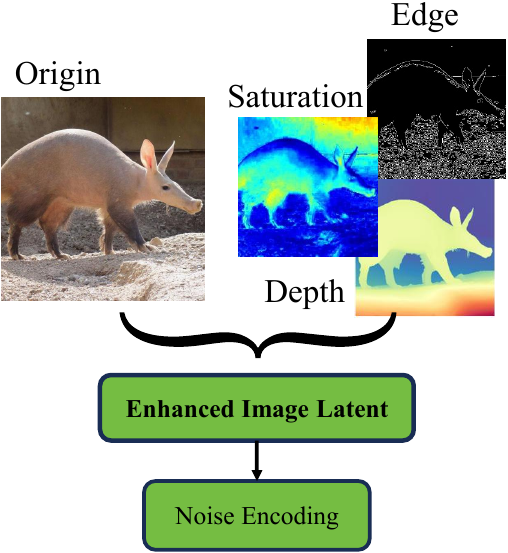}
    \caption{Image Augmentation Process. The edge detection map, saturation heatmap, and depth map were extracted in advance and then combined with the original image to form the augmented image set for each sample.}
    \label{fig:imgEnhance}
\end{figure}

\subsection{Joint-Modal Attention}
Most existing EEG-conditioned synthesis approaches adopt an Diffusion model backbone that injects EEG signals through cross-attention mechanisms. These frameworks commonly aligned EEG representations with graphical and textual latent space to ensure training stability. However, This design introduces structural limitations. Text-aligned EEG representations restrict the expressive cognition in EEG signals and degrade perceptual information that is not explicitly semantic. In addition, conventional cross-attention formulations constrain dual-interactions between modalities and limit adaptive information exchange \cite{MMDiT}.

To overcome these limitations, we propose a Joint-Modal Attention mechanism. The proposed formulation integrates image, text, and EEG features within a unified modeling process. Each modality preserves its intrinsic representational structure while participating in shared attention operations. This joint modeling strategy enables richer cross-modal interactions and facilitates the coordinated representation of semantic and perceptual information.

Let $H_m \in \mathbb{R}^{L_m \times D}$ denote the sequence of tokens for a specific modality $m \in \{\text{img}, \text{txt}, \text{eeg}\}$, where $L_m$ represents the token sequence length and $D$ is the embedding dimension. Since these modalities have fundamentally differences, forcing them to share transformer weights would restrict the model's ability to process their distinct features. Therefore, we employ modality-specific learnable parameters for layer normalization and linear scale and shift projections. This ensures that EEG signals are processed in their native feature space without forced alignment to text.

Inside a Joint-modal Attention block, we first normalize the inputs using adaptive modulation conditioned on the diffusion timestep $\tau$. Let $\tilde{H}_m$ denote the normalized features. We then project these sequences into query ($Q$), key ($K$), and value ($V$) representations:
\begin{equation}
Q_m = \tilde{H}_m W_Q^{(m)}, \quad K_m = \tilde{H}_m W_K^{(m)}, \quad V_m = \tilde{H}_m W_V^{(m)}
\end{equation}
where $W_{\{Q,K,V\}}^{(m)} \in \mathbb{R}^{D \times D}$ are learnable projection matrices distinct for each modality.

To enable interaction, we concatenate these sequences along the temporal dimension to form a unified joint sequence:
\begin{equation}
\begin{aligned}
Q_{\text{joint}} &= \text{Concat}[Q_{\text{img}}, Q_{\text{txt}}, Q_{\text{eeg}}] \in \mathbb{R}^{L_{\text{total}} \times D} \\
K_{\text{joint}} &= \text{Concat}[K_{\text{img}}, K_{\text{txt}}, K_{\text{eeg}}] \in \mathbb{R}^{L_{\text{total}} \times D} \\
V_{\text{joint}} &= \text{Concat}[V_{\text{img}}, V_{\text{txt}}, V_{\text{eeg}}] \in \mathbb{R}^{L_{\text{total}} \times D}
\end{aligned}
\end{equation}

We then perform a single joint self-attention operation on this joint sequence:
\begin{equation}
\text{Attention}(Q_{\text{joint}}, K_{\text{joint}}, V_{\text{joint}}) = \text{softmax}\left(\frac{Q_{\text{joint}} K_{\text{joint}}^\top}{\sqrt{D}}\right) V_{\text{joint}}
\end{equation}

This operation facilitates a bi-directional information flow. Image tokens can attend to specific visual details in EEG tokens that are absent in text, while text and EEG tokens can interact to contextually refine noisy signals with high-level semantic guidance.

Following joint-attention layer, the output sequence is split back into its constituent modalities. We apply separate Multi-Layer Perceptrons (MLP) to each stream to compute the residuals:
\begin{equation}
H_m' = H_m + \text{MLP}_m\left( \text{Split}_m(\text{Attention}(\cdot)) \right)
\end{equation}

By employing modality-specific projection parameters while performing attention over a unified latent space, JMVR establishes a direct mapping from EEG representations to high-dimensional image features. This design avoids forcing EEG signals into a text-aligned latent space and thereby preserves perceptually relevant information related to spatial structure and chromatic characteristics.

% 3.4 扩散步骤相关门控
\subsection{Diffusion Step Gating}
High-fidelity visual reconstruction requires the model to explicitly capture the evolving dependency between semantic guidance and EEG signals throughout the generative process. Conventional conditioning strategies implicitly assume a temporally static contribution of each modality across all diffusion steps. Such an assumption is inconsistent with the coarse and fine nature of diffusion-based generation. In practice, high-level textual semantics mainly control global structure formation during early denoising stages under high noise conditions \( \tau \to \tau_{\max} \) and fine-grained EEG representations become increasingly critical for refining perceptual details at low noise regimes \( 0 \to \tau \).

To facilitate the temporal balanced modeling in joint-modal information integration, we introduce a Diffusion Step Gating strategy that explicitly regulates the information flow across diffusion timesteps. Building upon the adaLN-zero inherent to the DiT architecture \cite{DiT}, this approach implements a hybrid gating mechanism to modulate the information flow based on the generative stage in the residual pathways.

Let $\Delta H_m$ denote the output of the Joint Self-Attention module for modality $m$. The modulation of the gating strategy is formulated as:
\begin{equation}\label{eq:residual_gate}
H_m' = H_m + \left[ \lambda_m(\tau) \cdot \alpha_m(\tau_{\text{emb}}) \right] \cdot \Delta H_m
\end{equation}
where $\alpha_m(\tau_{\text{emb}})$ is a learnable coefficient to inject information from condition latent embedding \cite{DiT}. $\lambda_m(\tau)$ is a specialized timestep prior designed to constraint information emphasis for text and EEG modalities.

Following the coarse and fine generative dynamics, we define $\lambda_m(\tau)$ as the timestep prior using complementary sinusoidal schedules:
\begin{align}
\lambda_{\text{txt}}(\tau) &= \sin \left( \frac{\tau}{T_{\max}} \cdot \frac{\pi}{2} \right) \\
\lambda_{\text{eeg}}(\tau) &= 1 - \sin \left( \frac{\tau}{T_{\max}} \cdot \frac{\pi}{2} \right)
\end{align}

By effectively coordinating the high-level textual semantics with fine-grained EEG visual cognition, the proposed diffusion step gating strategy promotes the decoupling utilization of joint-modal information, enhancing the fidelity of the reconstructed visual stimuli with precise cognitive details.

\begin{figure*}[h]
    \centering
    \includegraphics[width=0.9\linewidth]{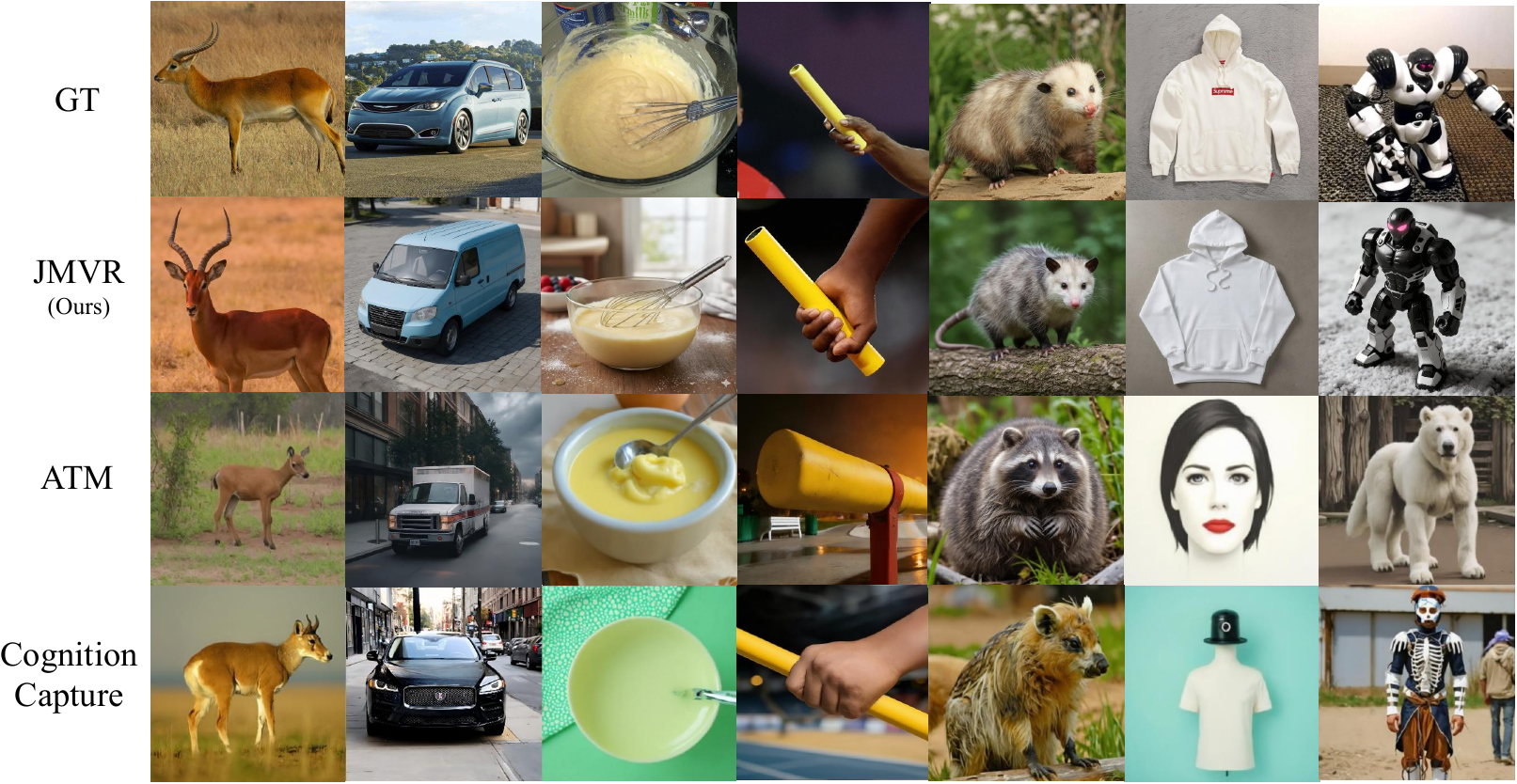}
    \caption{Visual comparison of reconstruction quality (The reconstruction samples were selected from subject-08).}
    \label{fig:picture}
\end{figure*}

\begin{table*}[h]
\centering
\setlength{\tabcolsep}{4.5pt}
\begin{tabular}{@{}lcccccccccc@{}} 
\toprule
\multicolumn{1}{c}{Method} & \multicolumn{5}{c}{Fine-grained} & \multicolumn{4}{c}{Coarse-grained} \\ 
\cmidrule(lr){2-6} \cmidrule(lr){7-10}  
& PixCorr $\uparrow$ & SSIM $\uparrow$ & LabEMD $\downarrow$ & DeepEMD $\downarrow$ & Alex(2) $\uparrow$ & Alex(5) $\uparrow$ & Incep $\uparrow$ & CLIP $\uparrow$ & SwAV $\downarrow$ \\  
\midrule
NICE (2024) & 0.173 & 0.326 & 23.409 & 2.834 & 0.783 & 0.832 & 0.746 & 0.773 & 0.523 \\ 
MUSE (2024) & 0.199 & 0.358 & 21.930 & 2.501 & 0.802 & 0.853 & 0.754 & 0.793 & 0.513 \\ 
ATM (2024) & 0.182 & 0.353 & 23.441 & 3.040 & 0.776 & 0.866 & 0.755 & 0.790 & 0.545 \\
DreamDiffusion (2024) & 0.206 & 0.347 & 22.764 & 1.741 & 0.787 & 0.857 & 0.751 & 0.791 & 0.504 \\ 
Perceptogram (2025) & 0.214 & 0.334 & 19.899 & 1.659 & \underline{0.856} & 0.874 & 0.762 & \underline{0.818} & 0.531\\
CognitionCapturer (2025) & 0.178 & 0.359 & 17.749 & 2.435 & 0.806 & \underline{0.894} & 0.769 & 0.803 & 0.552 \\
JMVR* (ours) & \underline{0.215} & \underline{0.367} & \underline{15.263} & \underline{1.123} & 0.821 & 0.877 & \underline{0.778} & 0.809 & \underline{0.493} \\ 
JMVR (ours) & \textbf{0.236} & \textbf{0.372} & \textbf{12.413} & \textbf{0.982} & \textbf{0.874} & \textbf{0.920} & \textbf{0.793} & \textbf{0.829} & \textbf{0.458} \\ 
\bottomrule
\end{tabular}
\caption{Quantitative assessments of the mean reconstruction quality of all subjects. The \textbf{best} and \underline{suboptimal} results are denoted by bold and underline respectively. *indicate that the text input was replaced with aligned EEG signals so that JMVR* became a EEG-to-image model.} 
\label{tab:quanti}
\end{table*}

\section{Dataset and Experiment Setup}

As one of the most commonly used datasets, ThingsEEG dataset is employed in this study which records EEG responses from 10 participants under a rapid serial visual presentation (RSVP) paradigm \cite{Things-EEG}. Visual stimuli are drawn from the THINGS database and consist of 16,740 natural images spanning a wide range of object categories \cite{Things}. The dataset is split into 1,654 training categories containing 10 images for each and 200 test categories containing only single image for each. 

Preprocessing includes segmenting the EEG recordings from 0 to 1000 ms for each stimulus, filtering between 0.1 and 100 HZ , and normalized. The baseline correction is performed using the 200ms EEG interval before stimulus. We employ Qwen3-VL-Embedding-2B for image detailed captioning and adopt CLIP-L/G for fine-grained and coarse-grained text encoding. The proposed method is conducted on double NVIDIA A40 48GB GPUs. Detailed hyperparameter settings are provided in our code repository.

Following the outline in CognitionCapture, we use Pixel Correlation (PixCorr) and Structural Similarity Index (SSIM) as pixel-level metrics. Additionally, we implement deep learning-based metrics including AlexNet (Layers 2 and 5), Inception, CLIP, and SwAV to evaluate semantic similarity \cite{cognitioncapturer}. To further evaluate the model's capability in reconstructing the color and spatial information within EEG signals, we introduce two novel metrics: CIELab EMD (LabEMD) and Depth EMD (DeepEMD). LabEMD is designed to evaluate color reconstruction consistency with human visual perception. It presents luminance and chromaticity differences between reconstructed image and original stimulus by calculating the Earth Mover's Distance (EMD) within the CIELab color space. DeepEMD aims to assess the preservation of spatial layouts. We implemented a commonly used deep analysis model called Depth-Anything-v2 to extract depth maps \cite{byung2025jointdit,depth_anything_v2}. Then, the EMD between these maps is computed to quantify the accuracy of the reconstructed spatial structures.

\begin{table*}
\centering
\setlength{\tabcolsep}{4pt}
\begin{tabular}{@{}lcccccccccc@{}} 
\toprule
\multicolumn{1}{c}{Method} & \multicolumn{5}{c}{Fine-grained} & \multicolumn{4}{c}{Coarse-grained} \\ 
\cmidrule(lr){2-6} \cmidrule(lr){7-10}  
& PixCorr $\uparrow$ & SSIM $\uparrow$ & LabEMD $\downarrow$ & DeepEMD $\downarrow$ & Alex(2) $\uparrow$ & Alex(5) $\uparrow$ & Incep $\uparrow$ & CLIP $\uparrow$ & SwAV $\downarrow$ \\  
\midrule
w/o Diffusion Step Gating & 0.228 & 0.363 & 12.633 & 0.985 & 0.865 & 0.885 & 0.790 & 0.814 & 0.527 \\ 
w/o Multi-Scale EEG & 0.223 & 0.364 & 13.172 & 1.065 & 0.858 & 0.877 & 0.788 & 0.816 & 0.551 \\ 
w/o Image Augmentation & 0.213 & 0.347 & 22.218 & 2.497 & 0.788 & 0.881 & 0.785 & 0.817 & 0.546 \\ 
JMVR*(our) & 0.215 & 0.367 & 15.263 & 1.123 & 0.821 & 0.877 & 0.778 & 0.809 & 0.493 \\ 
JMVR(our) & \textbf{0.236} & \textbf{0.372} & \textbf{12.413} & \textbf{0.982} & \textbf{0.874} & \textbf{0.920} & \textbf{0.793} & \textbf{0.829} & \textbf{0.458} \\ 
\bottomrule
\end{tabular}
\caption{Quantitative assessments of the mean reconstruction quality of the JMVR method in ablation study across all subjects. *indicate that the text input was replaced with aligned EEG signals so that JMVR* became a EEG-to-image model.} 
\label{tab:abl}
\end{table*}

\section{Experimental Results}

\subsection{Performance Comparison}

Tab. \ref{tab:quanti} shows the quantitative assessments of the image reconstruction quality. The results demonstrates that our JMVR framework achieves state-of-the-art performance across all evaluated metrics. Compared with alignment-centric multi-modal methods, joint-modal training demonstrates superior performance in image reconstruction and stronger multi-modal comprehension capabilities. This provides a novel training paradigm for visual reconstruction from neural signals. Although it is a fair comparison between JMVR with joint inputs and the methods that obtains text-like representations via EEG-text pre-alignment, the further study for the image reconstruction performance is conducted through JMVR* model that only uses EEG inputs without pre-alignment. The result shows that the restricted paradigm also superior to competing approaches in fine-grained reconstruction and maintains highly competitive capabilities in high-level semantic consistency.

Fig. \ref{fig:picture} provides the reconstruction results of JMVR comparing to the origin visual stimuli ground truth (GT) and competing approaches (ATM and CognitionCapturer). As illustrated, our method demonstrates a superior capability in reconstructing both high-level semantic consistency and low-level visual attributes. In contrast to the semantic drift effect in competing approaches, JMVR consistently captures the correct core category of the visual stimuli. This robustness validates the effectiveness the advanced semantic-understanding performance of learning text and EEG in a shared latent space. Furthermore, JMVR successfully reconstructs fine-grained details that are lost in other approaches like the color of minivan and the material of mixing bowl. This further illustrates the contribution of undegraded joint-model framework and multi-scale EEG feature extraction method.

\begin{figure}
    \raggedright
    \includegraphics[width=0.8\linewidth]{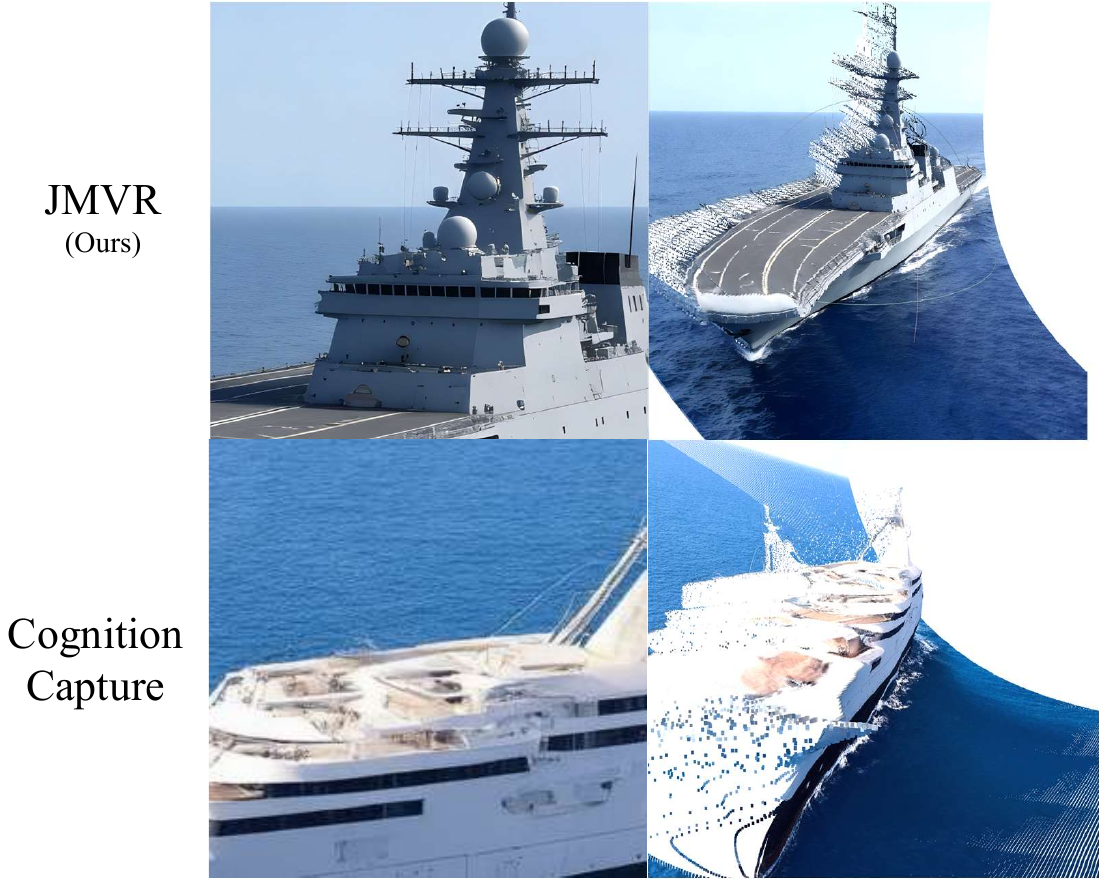}
    \caption{3D point clouds of the reconstruction image (The sample was selected from the aircraft carrier reconstruction by Subject-08).}
    \label{fig:3d}
\end{figure}

\subsection{Ablation Experiments}

Tab. \ref{tab:abl} presents the individual contributions of our proposed components. The ablation of the multi-scale EEG feature extraction and the diffusion step gating leads to a consistent performance decrease across all metrics. This decline highlights the significance of these components for EEG signal encoding and the joint-modal flow matching regulation. Furthermore, the exclusion of image enhancement module causes a sharp increase of LabEMD and DeepEMD, demonstrating its indispensability for fine-grained visual fidelity.

\section{Discussion}

\subsection{Spatial Reconstruction Analysis}

Since DeepEMD in Tab. \ref{tab:quanti} are often difficult to capture by human eye, we implemented the Depth-Anything-v2 model to project the reconstructed images into 3D point clouds for further discussion. Fig. \ref{fig:3d} demonstrates the spatial reconstruction performance of two models that incorporate depth information during training. Although the Cognition Capture model pre-aligns EEG signals with depth information during training, its strategy appears to limited spatial comprehension. The upper structure of the ship fails to be accurately reconstructed. In contrast, JMVR relatively reconstructs correct depth hierarchical relationships with the joint-training approach. The results demonstrates that our framework more effectively captures the latent interplay between EEG-encoded visual perception and the spatial semantics of images.

\subsection{Interaction between Text and EEG Modalities}

\begin{figure}
    \centering
    \begin{subfigure}{0.23\textwidth}
        \raggedright
        \includegraphics[width=\linewidth]{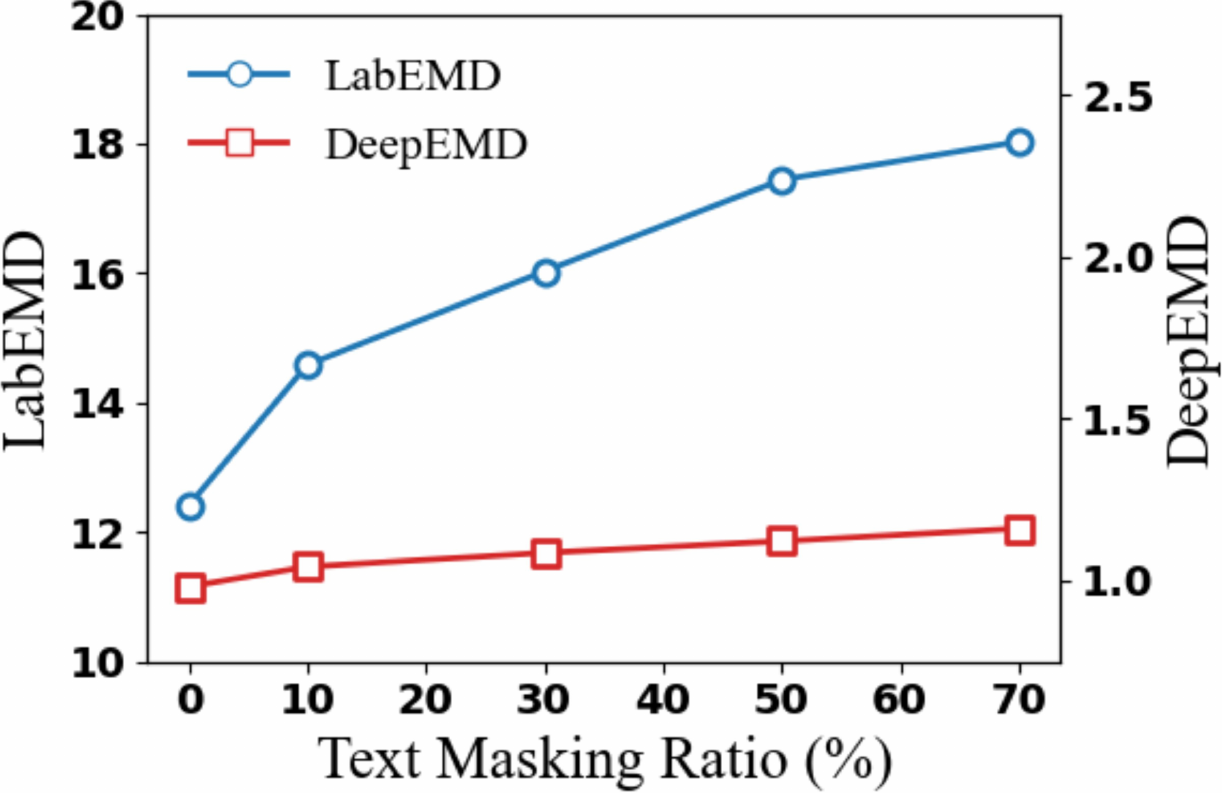}
        \caption{Text masking}
        \label{subfig:textMask}
    \end{subfigure}
    \hfill
    \begin{subfigure}{0.23\textwidth}
        \raggedright
        \includegraphics[width=\linewidth]{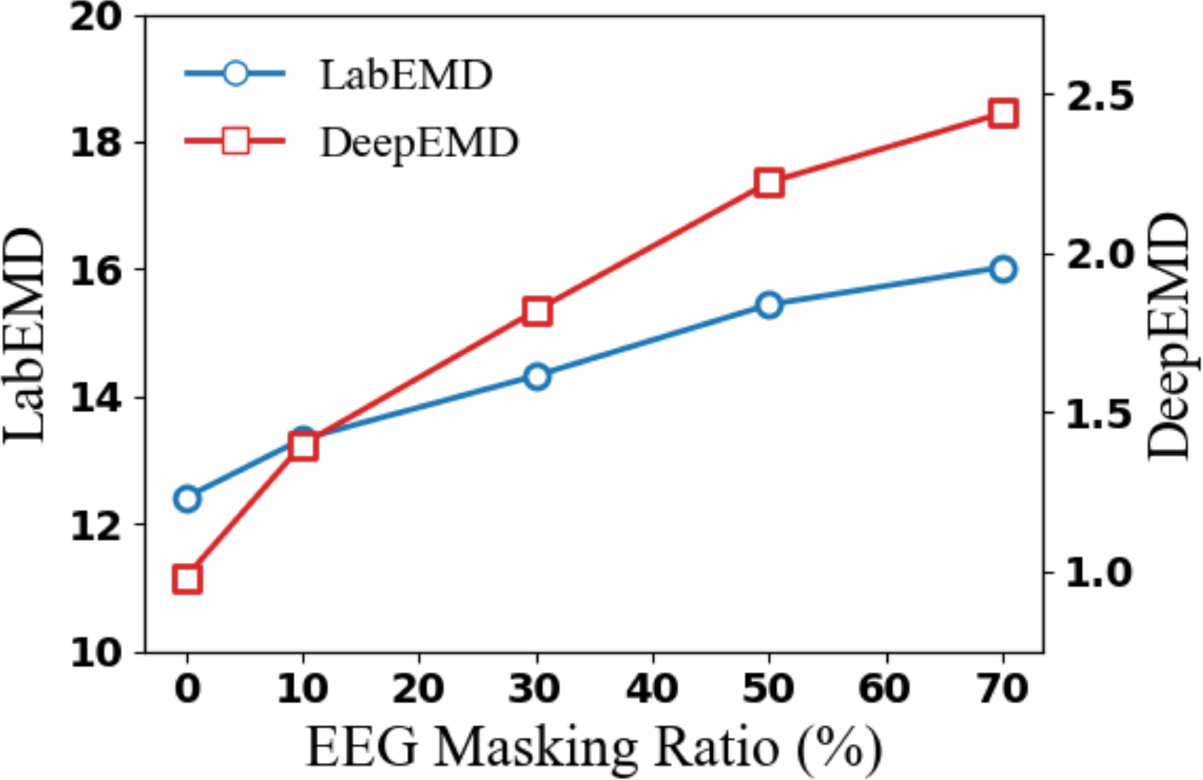}
        \caption{EEG masking}
        \label{subfig:EEGMask}
    \end{subfigure}
    \caption{LabEMD and DeepEMD performance under different text and EEG masking ratios.}
    \label{fig:masking}
\end{figure}

To verify the respective significance of text and EEG modalities in visual reconstruction, we analyzed the variations in LabEMD and DeepEMD metrics under different masking ratios. Since text encodings typically provide explicit information of object coloration. Color fidelity declines as the text masking ratio increases as shown in Fig. \ref{subfig:textMask}. When the text masking ratio becomes higher, the color reconstruction capability of JMVR degrades to our comparison EEG-to-image model. The results underscores the necessity of incorporating joint-modal training with textual guidance for high-quality visual reconstruction.
Conversely, Fig. \ref{subfig:EEGMask} reveals that EEG input masking triggers a simultaneous increase in both LabEMD and DeepEMD metrics. This indicates that EEG signals containing rich visual attributes related color and spatial depth. These can effectively complement semantic information of the textual modality. Thus, the result highlights a limitation of comparative methods that inevitably discard these perceptual details inherent in the neural signal by aligning EEG directly with text embeddings, indicating the advantages of joint-modal paradigm.

\begin{figure}
    \centering
    \begin{subfigure}{0.23\textwidth}
        \centering
        \includegraphics[width=\linewidth]{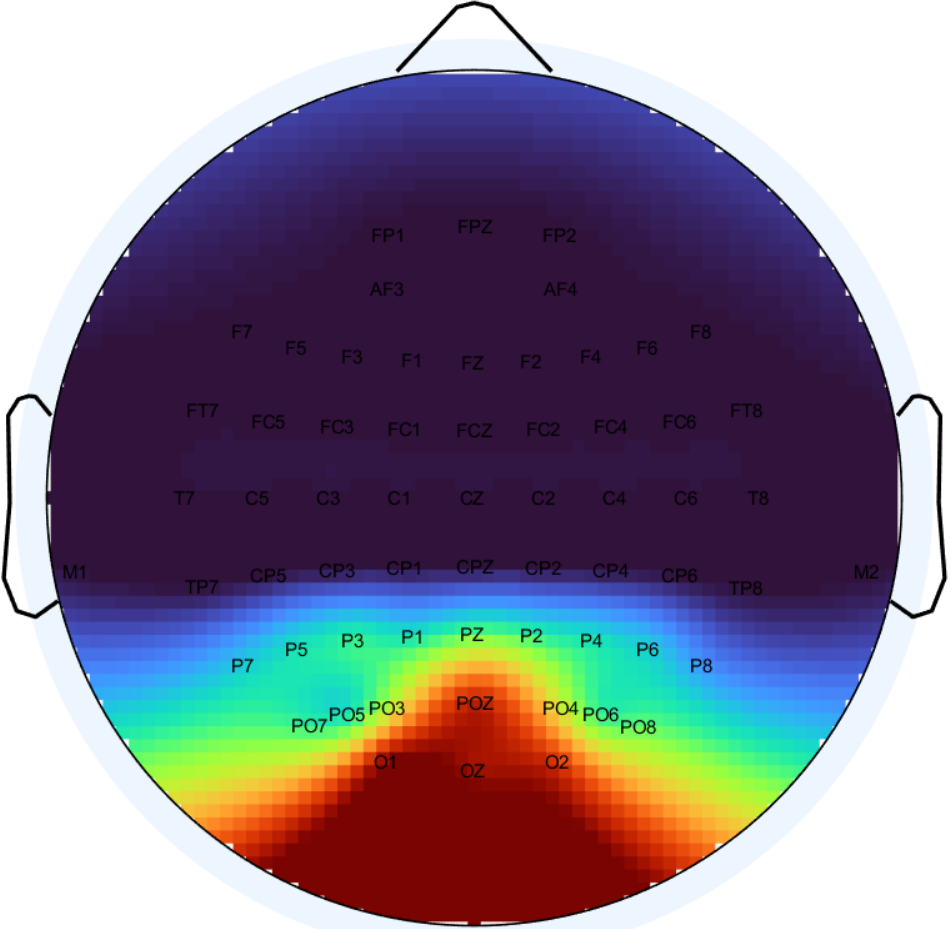}
        \caption{single-modal \\ (EEG-only)}
    \end{subfigure}
    \hfill
    \begin{subfigure}{0.23\textwidth}
        \centering
        \includegraphics[width=\linewidth]{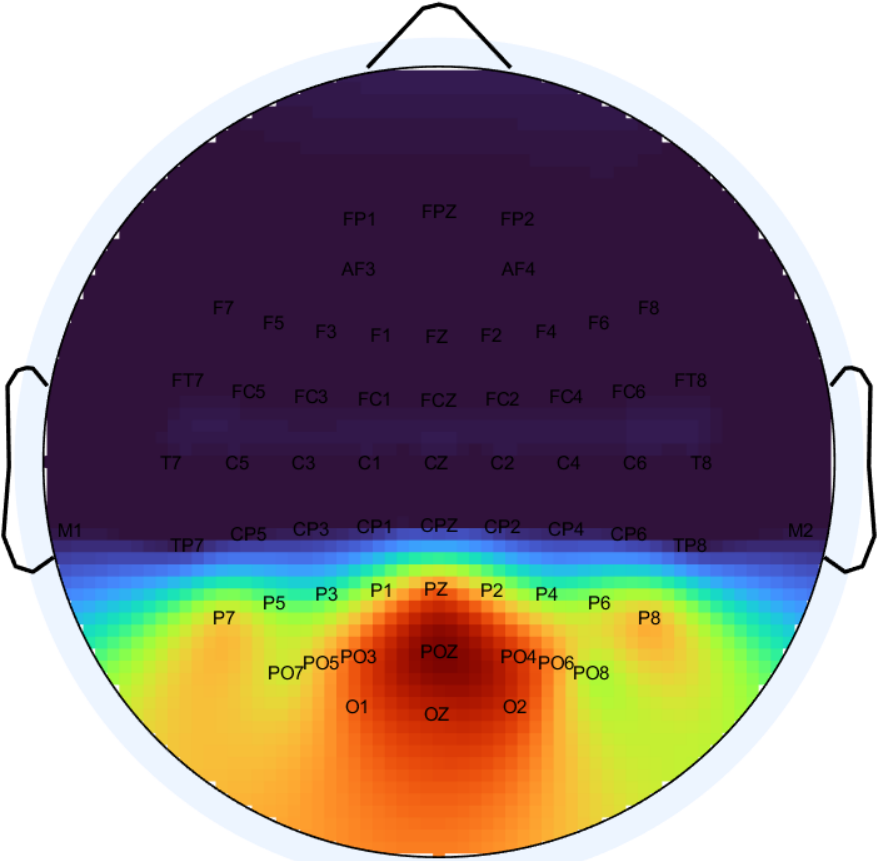}
        \caption{joint-modal \\(EEG $+$ text)}
    \end{subfigure}
    \caption{Channel weights of single-modal JMVR* with only EEG input and joint-modal JMVR with both EEG and text inputs.}
    \label{fig:weight}
\end{figure}

To further examine how textual semantic information modulates the internal mechanisms of EEG-based neural decoding models, we compared the spatial distributions of channel attention weights under single-modal and multi-modal conditions in Fig. \ref{fig:weight}. 
%The visualization results in  provide direct evidence of how multi-modal inputs alter the model’s reliance on neural information.

Under the single-modal condition, the attention weights were highly concentrated on occipital electrodes (e.g., OZ, O1, O2, and POZ), which are primarily associated with low-level visual feature encoding \cite{occipital}. This illustrates of spatial concentration reflects an information bottleneck in single-modality EEG decoding. In the absence of high-level semantic guidance, the model is forced to rely on channels with direct stimulus-related responses in order to extract visual information.

With the introduction of textual descriptions, the distribution of attention weights was systematically reorganized. Although occipital regions remained prominent, attention extended toward parietal and temporo-occipital areas. From a perspective of neural functions, these regions are known to support spatial attention, object integration, and visual–semantic mapping \cite{brain1,brain2}. From a modeling perspective, the availability of a stable global semantic from text reduces the burden on EEG signals to encode object identity and allows attention to be redistributed toward channels with spatial structure and fine-grained perceptual details. This redistribution supports more precise image reconstruction from EEG by preserving sensitivity to visual cognition attributes.

Overall, the observed reorganization of channel attention reveals an interactive mechanism between textual and EEG modalities within the joint attention space. By introducing high-level semantic constraints, multi-modal learning shifts EEG encoding from early visual cortex toward a cooperation of neural signals across multiple brain regions. These findings offer mechanistic insight into joint-modal neural decoding and suggest a principled pathway for improving EEG-based visual reconstruction through joint-cognitive inputs.

\subsection{Temporal Evolution of Visual Cognition}

\begin{figure}
    \centering
    \includegraphics[width=0.8\linewidth]{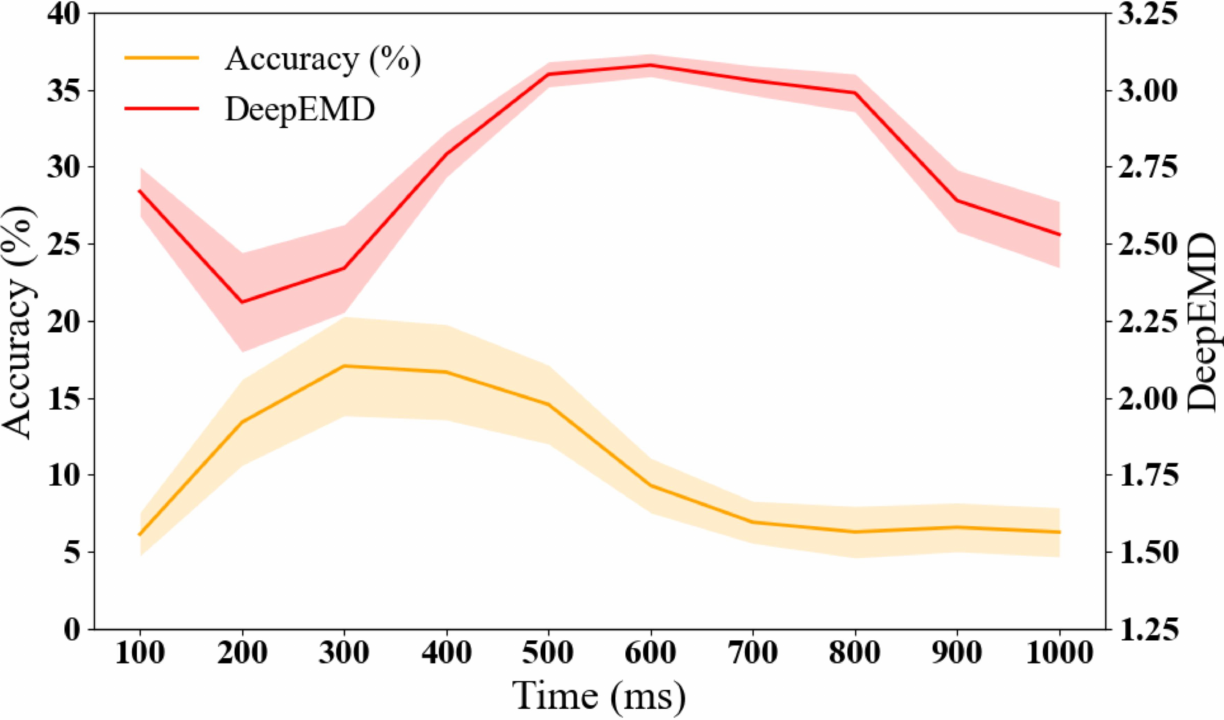}
    \caption{Temporal evolution of semantic classification (Accuracy) and structural reconstruction quality (DeepEMD). Note that a decrease in the DeepEMD metric corresponds to an improvement in spatial reconstruction performance. The double valleys in the figure represent two peaks in spatial reconstruction performance.}
    \label{fig:timedetial}
\end{figure}

To investigate the focus of visual cognition, we analyzed the performance of the unnimodal model JMVR* driven only by EEG signals within different time interval. The EEG data was segmented using a sliding window of $[t-100,t]$ ms for decoding. The reconstructed images being classified via 10-way zero-shot classification with the CLIP model. Since the sparsity of the segmented EEG information causes high variance for CIELab EMD, our analysis only focuses on semantic categorization and depth reconstruction. As illustrated in Fig. \ref{fig:timedetial}, the semantic classification accuracy exhibits a unimodal profile, peaking within the 200–300 ms interval. This suggests that subject identification is primarily an early-stage process occurring shortly after stimulus. 
In contrast, the DeepEMD metric reveals a distinct bimodal distribution. During the early 100–300 ms phase , DeepEMD values are significantly lower than the baseline. This indicates an early spatial understanding even before the peak of semantic classification cognition. The early structural encoding may be linked to mechanisms of spatial localization or figure-ground segregation within the visual pathway. In the later stage of visual cognition during 900–1000 ms, we observe a resurgence in spatial reconstruction quality. The results demonstrate that after the initial semantic recognition, the brain re-focused on depth information and spatial details to refine its cognitive representation of the scene.

\section{Conclusion}
To mitigate the loss of cognitive information in EEG-based visual reconstruction, we propose the Joint-Modal Visual Reconstruction framework. The model incorporates four key components: multi-scale EEG encoding, image augmentation, joint-modal attention, and diffusion step gating. It achieves state-of-the-art performance in reconstructing visual stimuli. Beyond performance improvements, we analyze the interaction mechanisms between multi-modal representations, offering new insights into the neural basis of visual reconstruction. Furthermore, we investigate the temporal dynamics of visual cognition, revealing how depth information and semantic subject features are cognized across different stages of neural processing. Despite these advances, the current evaluation process mainly emphasizes depth consistency and chromatic fidelity and lacks a comprehensive assessment of higher-level visual cognitive reconstruction. Future work may extend the framework to incorporate additional modalities and develop more informative evaluation metrics to quantify visual cognitive reconstruction performance.

\newpage

\bibliographystyle{named}
\bibliography{ijcai_26}

\end{document}